\documentclass[conference]{IEEEtran}

\usepackage{times}
\usepackage{soul}
\usepackage{url}
\usepackage[hidelinks]{hyperref}
\usepackage[utf8]{inputenc}

\usepackage{subcaption}
\usepackage{caption}
\usepackage{amsmath}
\usepackage{amsthm}
\usepackage{booktabs}
\usepackage{algorithm}
\usepackage{algorithmic}
\usepackage[switch]{lineno}
 \usepackage{amssymb}
\usepackage{graphicx}
\usepackage{amsmath}
\usepackage{bbm}

\usepackage{color}
\usepackage[normalem]{ulem}
\usepackage{rotating}
\usepackage{tabularray}
\usepackage{amssymb}

\IEEEoverridecommandlockouts
\usepackage{cite}
\usepackage{amsmath,amssymb,amsfonts}
\usepackage{algorithmic}
\usepackage{graphicx}
\usepackage{textcomp}
\usepackage{xcolor}
\def\BibTeX{{\rm B\kern-.05em{\sc i\kern-.025em b}\kern-.08em
    T\kern-.1667em\lower.7ex\hbox{E}\kern-.125emX}}
    
\newlength\myindent
\linenumbers

\title{Anomaly Prediction: A Novel Approach with Explicit Delay and Horizon}

\author{
Jiang YOU$^{1,2,3}$\and
René NATOWICZ$^{1, 2}$\and
Arben CELA$^{1, 2}$\and
Jacob OUANOUNOU$^{3}$\and
Patrick SIARRY$^{1}$\\
\affiliations
$^1$Laboratoire Images, Signaux et Systèmes Intelligents (LISSI),  Université Paris-Est Créteil\\
$^2$Département Informatique et Télécommunication, ESIEE Paris-Université Gustave Eiffel\\
$^3$HN-Services, Paris\\
\emails
\{jiang.you, rene.natowicz, arben.cela\}@esiee.fr, 
jouanounou@hn-services.com, 
siarry@u-pec.fr
}

\begin{document}
\nolinenumbers

\title{Anomaly Prediction: A Novel Approach with \\Explicit Delay and Horizon \\

\thanks{
{
\textsuperscript{1} Laboratoire Images, Signaux et Systèmes Intelligents (LISSI), Université Paris-Est Créteil , Île-de-France, France\\\textsuperscript{2} Département Informatique et Télécommunication, ESIEE Paris-Université Gustave Eiffel, Île-de-France, France
 \\\textsuperscript{3} HN-Services, Île-de-France, France
 \\\textsuperscript{4} AI Laboratory, University Metropolitan Tirana, Tirana, Albania } 
 \vspace{0.5em}\\
 \hspace*{0.2em} \textit{  Proceedings of the 20\textsuperscript{th} International Conference on Intelligent Computer Communication and Processing}, October 2024, Cluj-Napoca, Romania.  
  \vspace{0.5em}\\
 979-8-3315-3997-9/24/\$31.00 ©2024 European Union
}

}

\author{

\hspace{5.0ex}
\and 

\IEEEauthorblockN{ Jiang YOU  \textsuperscript{1, 2, 3} }
\IEEEauthorblockA{
\textit{Université Paris-Est Créteil, }\\
Île-de-France, France\\
jiang.you@esiee.fr}
\and
\hspace{10.0ex}
\and 
\IEEEauthorblockN{Arben CELA \textsuperscript{1, 2, 4} }
\IEEEauthorblockA{
\textit{ESIEE Paris-UGE}\\
Île-de-France, France\\
arben.cela@esiee.fr}
\and
\hspace{10.0ex}
\and 
\IEEEauthorblockN{René NATOWICZ \textsuperscript{1,2} }
\IEEEauthorblockA{
\textit{ESIEE Paris-UGE}\\
Île-de-France, France\\
rene.natowicz@esiee.fr}
\and 
\hspace{13.0ex}
\and
\hspace{2.0ex}
\and
\IEEEauthorblockN{Jacob OUANOUNOU \textsuperscript{3} }

\IEEEauthorblockA{
\textit{HN-Services}\\
Île-de-France, France\\
jouanounou@hn-services.com}
\and
\IEEEauthorblockN{Patrick SIARRY \textsuperscript{1} }
\IEEEauthorblockA{
\textit{Université Paris-Est Créteil}\\
Île-de-France, France\\
siarry@u-pec.fr}
}

\maketitle

\begin{abstract}
Anomaly detection in time series data is a critical challenge across various domains. Traditional methods typically focus on identifying anomalies in immediate subsequent steps, often underestimating the significance of temporal dynamics such as delay time and horizons of anomalies, which generally require extensive post-analysis. This paper introduces a novel approach for time series anomaly prediction, incorporating temporal information directly into the prediction results. We propose a new dataset specifically designed to evaluate this approach and conduct comprehensive experiments using several state-of-the-art methods. Our results demonstrate the efficacy of our approach in providing timely and accurate anomaly predictions, setting a new benchmark for future research in this field.
\end{abstract}

\section{Introduction}
Time series anomaly detection has a broad application across various domains, reflecting its critical importance and versatility. In finance, it is used to detect fraudulent transactions \cite{time_series_anomaly_detection_finance_2006} and unusual market activities \cite{time_series_anomaly_detection_financemarket_activity_2015}. In healthcare, it aids in monitoring patient vital signs and detecting early signs of medical conditions \cite{ecg_heart_anomaly_detection_2007}. Industrial applications include predictive maintenance, where detecting anomalies in machine operations can prevent equipment failures \cite{li_equipement_failure_deep_2019}. In server reliability engineering, anomaly detection is essential for identifying unexpected server breakdowns \cite{anomaly_detection_microsoft_2019}. 

Over time, methodologies in time series anomaly detection evolved significantly \cite{hameurlain_anomaly_survey_time_series_2021}. Initially, the field was dominated by statistical methods and simple machine learning techniques such as Autoregressive Moving Average (ARMA) \cite{arma_anomaly_detection}, K-Means Clustering \cite{kmeans_anomalydetection_2007_TrafficAD}, Matrix Profile \cite{matrix_proffile_anomaly_time_series_2020} etc. However, with increasing data complexity, sophisticated models like deep learning approaches such as OmniAnomaly \cite{omnianomaly_2019}, TranAD \cite{tranad_anomaly_2022}, Anomaly Transformer \cite{Xu2021AnomalyTransformer} have become mainstream.

While the classical approach of anomaly detection is to identify anomalies as they occur or in the immediate future \cite{nab_lavin_evaluating_2015} \cite{laptev_yahoo_dataset_2017},  recent research has been interested in providing alerts with sufficient lead times to enhance detection capabilities \cite{Lead_Time_Analysis_Failure_Prediction_2023} \cite{cerqueira_early_anomaly_detection_2023}. Despite these advances, there are still significant difficulties in predicting the temporal dynamics of anomalies, particularly in predicting the lead time (or delay time) and precise range of predicted anomalies(or horizon), as well as the cumbersomeness of training models.

Recent critiques have also highlighted issues with current research methodologies, particularly concerning the datasets used. Many datasets do not adequately reflect real-world complexities, leading to an underestimation of deep learning methods' effectiveness \cite{current_time_series_are_flawed_2022}. In some cases, simpler models like ARMA outperform their complex counterparts (Long Short-term Memory (LSTM), Transformer), suggesting the need for more complicated tasks and datasets \cite{si2024timeseriesbench}.

\begin{table}[t!]
\centering
\setlength{\tabcolsep}{6.0pt} 
\caption{Comparison of Anomaly Detection and Anomaly Prediction Approach}
\begin{tabular}{l|c|c}
\toprule
\textbf{Approach} & \textbf{Anomaly Detection} & \textbf{Anomaly Prediction} \\ 
\midrule
\textbf{Inputs} & Time series features & Time series features \\ 
\midrule
\textbf{Outputs} & Scalar or binary value & Probability densities \\
\midrule
\begin{tabular}[c]{@{}c@{}} \textbf{Problems} \\ \textbf{Formulation} \end{tabular} & \begin{tabular}[c]{@{}c@{}}$f(x) : x \rightarrow y$ \\ $R^{L \times M} \rightarrow R$\end{tabular} & \begin{tabular}[c]{@{}c@{}}$f(x) : x \rightarrow y$ \\ $R^{L \times M} \rightarrow R^{T}$\end{tabular} \\ 
\midrule
\textbf{Methods} & \begin{tabular}[c]{@{}c@{}}Statistical techniques, \\ Classification techniques, \\ Clustering algorithms, \\ Reconstruction algorithms\end{tabular} & \begin{tabular}[c]{@{}c@{}}Multi-step Time Series \\ Forecasting algorithms\end{tabular} \\
\midrule
\begin{tabular}[c]{@{}c@{}}\textbf{Loss} \\ \textbf{Function}\end{tabular}
& \begin{tabular}[c]{@{}c@{}}Binary Cross-Entropy, \\ Mean Square Error, etc\end{tabular} & \begin{tabular}[c]{@{}c@{}}Wasserstein Loss \\ (Cumulative Sum)\end{tabular} \\ 
\midrule
\begin{tabular}[c]{@{}c@{}}\textbf{Evaluation} \\  \textbf{Metrics} \end{tabular} & \begin{tabular}[c]{@{}c@{}}Precision, Recall, \\ F1 Score, ROC-AUC, etc\end{tabular} & \begin{tabular}[c]{@{}c@{}}Existence, Density, \\ Lead Time, Dice Score\end{tabular} \\ 
\bottomrule
\end{tabular}
\label{table:anomaly_prediction_difference}
\end{table}

Recognizing these challenges, this paper introduces a novel approach called "Anomaly Prediction". Instead of merely predicting the occurrence of an anomaly or its severity, this approach predicts the distribution densities of anomaly events over future time intervals, providing clear delay time and horizon information. This enhances the practical utility of anomaly detection and demonstrates deep models' ability to recognize complex patterns in more difficult tasks.

To empirically validate our approach, we develop a synthetic dataset tailored to assess predictive models' performance in capturing complex temporal dynamics. We propose using Wasserstein loss to model the temporal dynamics by measuring the distribution densities of anomaly events. Additionally, to thoroughly evaluate the performance of models and understand the characters of the dataset, we introduce four additional metrics (Existence, Density, Lead Time, and Dice Score) for comprehensive performance evaluation (Table \ref{table:anomaly_prediction_difference}).

Through extensive experiments, we demonstrate that our method provides timely and precise predictions of anomalous events, integrating delay time and forecasting horizon. Our findings suggest that this approach substantially improves predictive models' utility in real-world settings, paving the way for future research in the field.

We summarize our contributions as follows:

\begin{itemize}
\item Introduction of a novel predictive approach for time series anomaly detection incorporating delay time and horizon.
\item Application of forecasting models and  Wasserstein loss to capture the distribution densities of anomaly events.
\item Development of a synthetic dataset and evaluation metrics to rigorously assess model performance.
\item Empirical validation of concept for our approaches.
\end{itemize}

In conclusion, this study pushes the boundaries of time series anomaly detection with the introduction of "Anomaly Prediction", a novel approach that integrates delay time and horizon information. 
 Future research will focus on refining these techniques and exploring their applicability in more complex scenarios, thereby expanding the scope and impact of anomaly detection in real-world applications.

\section{Related Works}
In this section, we provide a comprehensive review of anomaly detection models, ranging from traditional methods to contemporary deep learning methods. Next, we delve into the latest developments in time series forecasting methods. In addition, we introduce related tasks that intersect with anomaly detection, such as event prediction, human behavior detection, and explainable time series anomaly detection. Finally, we present anomaly prediction as an emerging field that combines elements of anomaly detection, time series forecasting, and event prediction.

\subsection{\textbf{Anomaly Detection}}
Supervised Anomaly Detection is similar to imbalanced binary classification. The challenges are generally from the discrimination to minority important samples and difficulty in collecting sufficient labels. \cite{anomaly_detection_time_series_review_2022}
 Unsupervised Anomaly Detection relieves these issues by taking advantage of them. The key idea of this approach is identifying the samples far from the majority or different from normal instances as anomaly. The mainstream techniques can be roughly divided into Forecasting Methods, Reconstruction Methods, Encoding Methods, Distance Methods, Distribution Methods, and Isolation Tree Methods \cite{anomaly_detection_time_series_review_2022}. 

Forecasting Methods involve creating a model to predict future values based on a current context window. These predicted values are compared to actual observed values to identify anomalies based on significant deviations. Representative algorithms include ARIMA \cite{hyndman2018forecasting} and LSTM-AD \cite{malhotra2015lstm}. 

Reconstruction Methods encode subsequences of normal training data into a low-dimensional latent space and then reconstruct them by expanding the latent vectors. Anomalies are detected by comparing the reconstructed subsequences to the original observed values, with discrepancies indicating potential anomalies. Notable examples of reconstruction methods are AutoEncoder (AE) \cite{sakurada2014anomaly}, and OmniAnomaly \cite{omnianomaly_2019}.

Encoding Methods also encode subsequences into a low-dimensional latent space, but unlike reconstruction methods, they compute the anomaly score directly from the latent space representations without attempting to reconstruct the original subsequences. This category includes algorithms such as GrammarViz \cite{senin2015time} and Series2Graph \cite{Boniol2020}.

Distance Methods utilize specialized distance metrics to compare points or subsequences within a time series. Anomalous subsequences are expected to have larger distances from clusters of normal subsequences \cite{Breunig2000}. 

Distribution Methods fit a distribution model with a given dataset and identify anomalies based on their probability or distance from the calculated distributions. Anomalies are typically found in the extremes or tails of the distributions, making this approach effective for identifying infrequent patterns. Key algorithms in this category are COPOD \cite{Li2020}, HBOS \cite{Goldstein2012}. 

Isolation Tree Methods construct an ensemble of random trees to partition the samples of a time series. Anomalies are detected based on the number of splits required to isolate a sample, with anomalous samples having shorter paths to the root of the tree \cite{Liu2008}. 

\subsection{\textbf{Time Series Forecasting}}
In time series forecasting, transformer-based models have significantly evolved to address the complexities of long-range dependencies and computational efficiency. Models like LogTrans \cite{li_2020_Enhancing}, Informer \cite{haoyietal-informer-2021} and PatchTST reduce computation complexity from $O(L\log(L)^2)$ to $O(\frac{L}{S}\log(\frac{L}{S}))$, where $S$ is the patch size, thereby enhancing the model's efficiency and ability to process long time series.

UNet-based methods, originally designed for image segmentation\cite{ronneberger_u-net_2015}, have also been adapted for time series forecasting due to their powerful feature extraction capabilities \cite{perslev_u-time_2019} \cite{Madhusudhanan_yformer_2023}. U-Net’s architecture, with its encoder-decoder structure and skip connections, facilitates the capture and merging of multi-scale information. Recently, Kernel U-Net \cite{you_kun_2024} has addressed specific challenges in time series forecasting by enhancing both expressiveness and computational efficiency. 

\subsection{\textbf{Event Prediction}}
In time series event prediction, methods are categorized into occurrence prediction, discrete-time prediction, continuous-time prediction, and point process \cite{event_prediction_2021}.

Occurrence prediction, the most straightforward, focuses on determining whether an event will occur within a future time period, often treated as a binary classification problem. Techniques for this include Support Vector Machines (SVMs) \cite{Inceoglu2018}, and decision trees\cite{DeCaigny2018}. Additionally, anomaly detection methods like one-class classification and hypothesis testing are used for rare event occurrences, leveraging the identification of deviations from normal samples. Regression-based approaches extend binary prediction to predict event count\cite{Ertugrul2018} or scale\cite{Gao2018}, enhancing the granularity of occurrence predictions.

Discrete-time prediction aims to forecast the approximate time slot (e.g., day, week, month) of an event's occurrence. This approach can be direct, where future time is partitioned into discrete values and predicted using regression or ordinal classification methods\cite{Tops2013}, or indirect, where time series forecasting techniques predict future feature values, and events are then identified within these predictions  \cite{Reid2018}. 

Continuous-time prediction, addressing the limitations of discrete-time methods, involves predicting the exact time of an event using techniques like regression, point processes, or survival analysis\cite{Simma2012}. Point processes model the conditional intensity function to predict event time values accurately \cite{Neumann2019FutureEP}, while survival analysis uses hazard functions to estimate event occurrence probabilities in the next time interval \cite{Qiao2018}. These methods provide high-resolution and precise event timing, adapting to complex temporal patterns in time series data.

\subsection{\textbf{Early recognition of ongoing human actions}}
Early recognition of ongoing human actions from video streams is a critical task with broad applications across various domains, including robotics, entertainment, surveillance, and healthcare \cite{WANGboyu_human_early_action_detection_201824}. The ability to promptly detect and recognize actions as they unfold is essential for enabling real-time decision-making and enhancing operational efficiency in these fields. Recent research \cite{WANGboyu_human_early_action_detection_201824} has proposed innovative methods utilizing Bidirectional Long Short-term Memory (BLSTM) networks to predict the temporal initiation of actions with high precision. These methods leverage advanced training techniques such as novel loss functions based on cumulative distribution functions, aiming to minimize prediction errors and improve overall recognition accuracy. 

\subsection{\textbf{Explainable Time Series Anomaly Detection}}
Exathlon is a benchmark specifically designed for explainable anomaly detection over high-dimensional time series data \cite{jacob2021exathlon}. Exathlon features real data traces derived from extensive executions of large-scale stream processing tasks on an Apache Spark cluster. They propose Metrics such as Anomaly Existence, Range Detection, Early Detection, and Exactly-Once Detection for unsupervised anomaly detection at instant. This benchmark enables the development and evaluation of a wide range of anomaly detection and explanation discovery techniques.

In this paper, we propose anomaly prediction as an emerging field that combines elements of anomaly detection, time series forecasting, and event prediction. This integrated approach not only identifies anomalies but also predicts their occurrence and assesses their potential impact, providing a comprehensive solution for proactive anomaly management. By leveraging the strengths of each component, anomaly prediction can offer earlier warnings and more accurate insights, enabling more effective responses to potential issues across various domains.

\section{Method}
In this section, we introduce our anomaly prediction approach, starting with problem formulation, where we define the task as predicting the distribution density of anomalies. We employ the Wasserstein distance as a loss function to measure the distribution similarity between actual anomalies and predicted data. To evaluate model performance, we use several metrics related to anomalies: Existence, Densities, Lead Time, and Dice Score. Our approach incorporates deep learning time series forecasting models to capture complex patterns and temporal shifts.

\begin{figure}[t!]
    \centering
    \includegraphics[width=\linewidth]{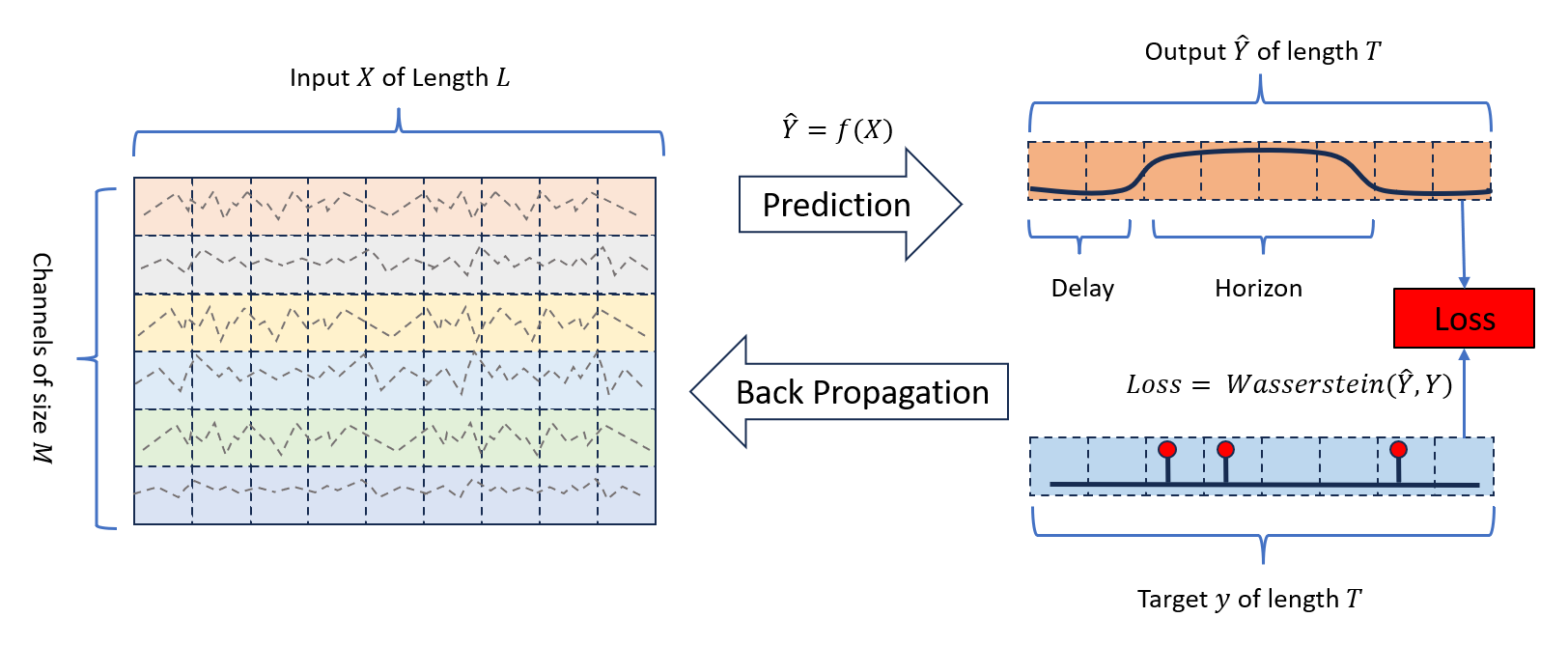}
    \caption{Illustration of  Anomaly Prediction Task}
    \label{fig:anomaly_prediction}
\end{figure}

\subsection{Problem Formulation}
\subsubsection{\textbf{Anomaly Prediction Task}}

Consider a multivariate time series dataset represented by the matrix $ x \in \mathbb{R}^{N \times M} $, where $ N $ represents the number of sampling times, and $ M $ represents the number of features. Let $ L $ be the length of the look-back window (memory), so the slice $(x_{t+1,1}, \dots, x_{t+L, M})$ (or, for short, $(x_{t+1}, \dots, x_{t+L})$ contains historical information about the system at instant $ t+1$. We note $(y_{t+L+1}, \dots, y_{t+L+T})$ the anomaly labels in the next $T$ step in the future horizon.   


The anomaly prediction problem can be formulated as:
\begin{align}
(\hat{y}_{t+L+1}, \dots, \hat{y}_{t+L+T}) = f(x_{t+1}, \dots, x_{t+L}) 
\end{align} 
where $ f $ is a function that predicts the probabilities $ \hat{Y}_{t+L+1} = (\hat{y}_{t+L+1}, \dots, \hat{y}_{t+L+T}) $ of finding anomalies based on the historical series $ (x_{t+1}, \dots, x_{t+L}) $. Here, $ \hat{y}_{t+L+i} $ represents the predicted overlapped probabilities of anomalies at time $t+L+i$ for $ i \in \{1, \dots, T\} $ (Figure \ref{fig:anomaly_prediction}).

Remark that classic time series forecasting task takes Mean Absolute Error (MAE) or Mean Square Error (MSE) as loss function. Instead of forecasting the actual value of time series, anomaly prediction predicts the densities of anomalous events. In this case, classical MAE or MSE loss is inappropriate because the presence and absence of anomaly events with different lead times contradict each other and make the model collapse. Therefore, We propose using Wasserstein loss, which integrates anomaly events densities over time dimension.

\subsubsection{\textbf{Wasserstein loss}}
Wasserstein loss (Cumulative sum loss) measures the distance or similarity of two distributions. We use it to evaluate the overall difference between predicted densities and actual Dirac distributions of anomaly labels over the entire future horizon. Technically, this method computes the MAE loss of the cumulative sum of predicted distributions and Dirac distributions over the time dimension. The Wasserstein loss is defined as:
   \begin{align}
   \text{Wasserstein}(\hat{Y}_t, Y_t) = \frac{2}{T(T+1)} \sum_{i=1}^{T} \left| \sum_{j=1}^{i} (\hat{y}_{t+j} - y_{t+j} ) \right|
   \end{align}
    where the MAE of cumulative sums between predicted values $\hat{y}$ and true values $y$ are compared at each time step $i$ within the prediction horizon. This loss function captures both the location and the shape of the predicted distribution relative to the actual distribution. It can approximate a mixture of up to $T$ distribution of anomalies in the prediction horizon. In some cases, the coefficient $\frac{2}{T+1}$ can be ignored to simplify the calculation.
    
\subsection{Metrics}
\label{metrics}
\begin{figure}[t!]
    \centering
    
    \begin{subfigure}[b]{\linewidth}
        \centering
        \includegraphics[width=\linewidth]{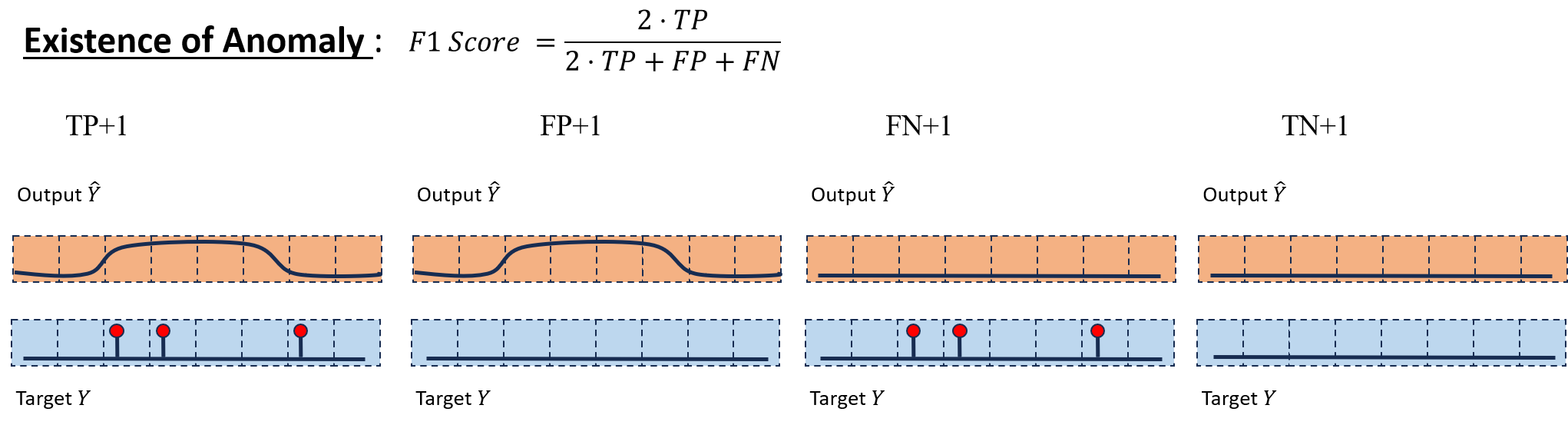}
        \caption{Existence of Anomaly}
        \label{fig:anomaly_existence}
    \end{subfigure}
    \vfill
    \begin{subfigure}[b]{\linewidth}
        \centering
        \includegraphics[width=\textwidth]{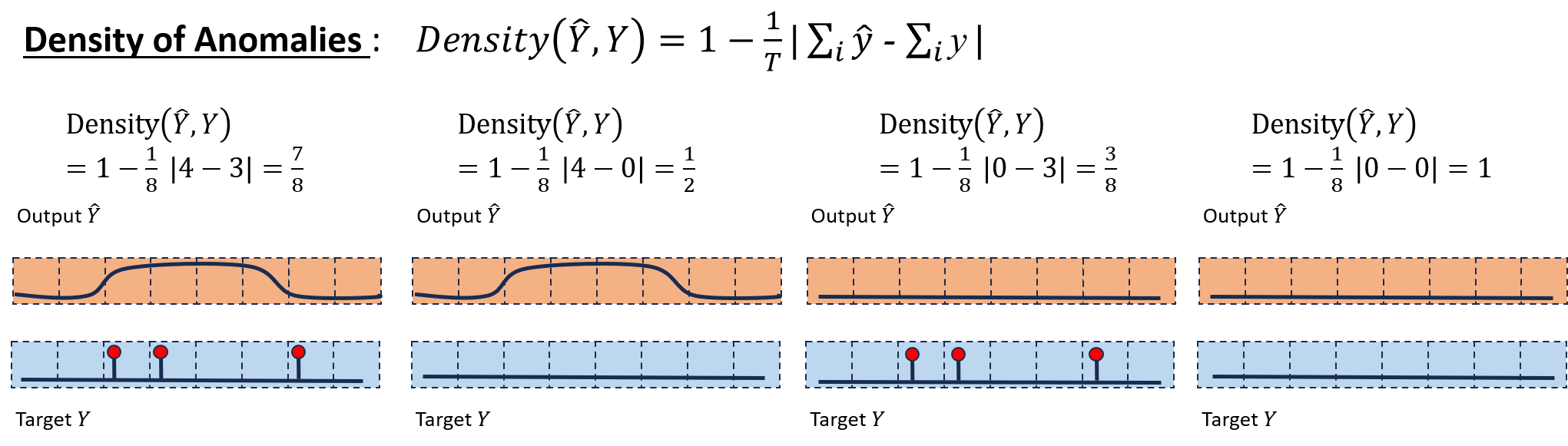}
        \caption{Density of Anomaly}
        \label{fig:anomaly_length}
    \end{subfigure}
    \vfill
    \begin{subfigure}[b]{\linewidth}
        \centering
        \includegraphics[width=\textwidth]{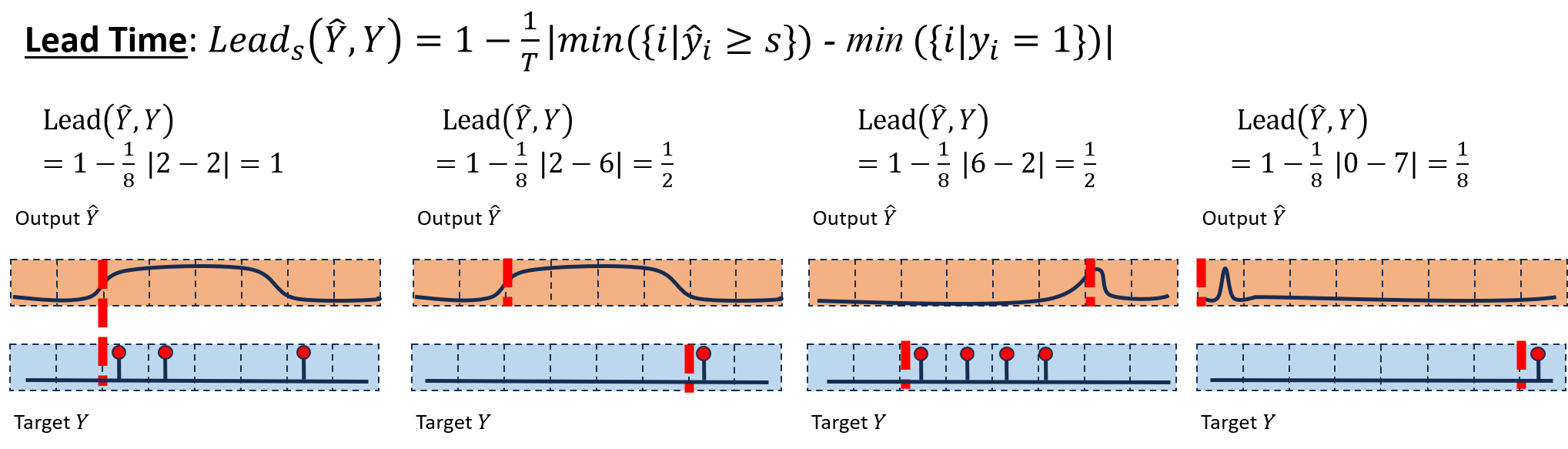}
        \caption{Lead Time}
        \label{fig:anomaly_leadtime}
    \end{subfigure}
    \vfill
    \begin{subfigure}[b]{\linewidth}
        \centering
        \includegraphics[width=\textwidth]{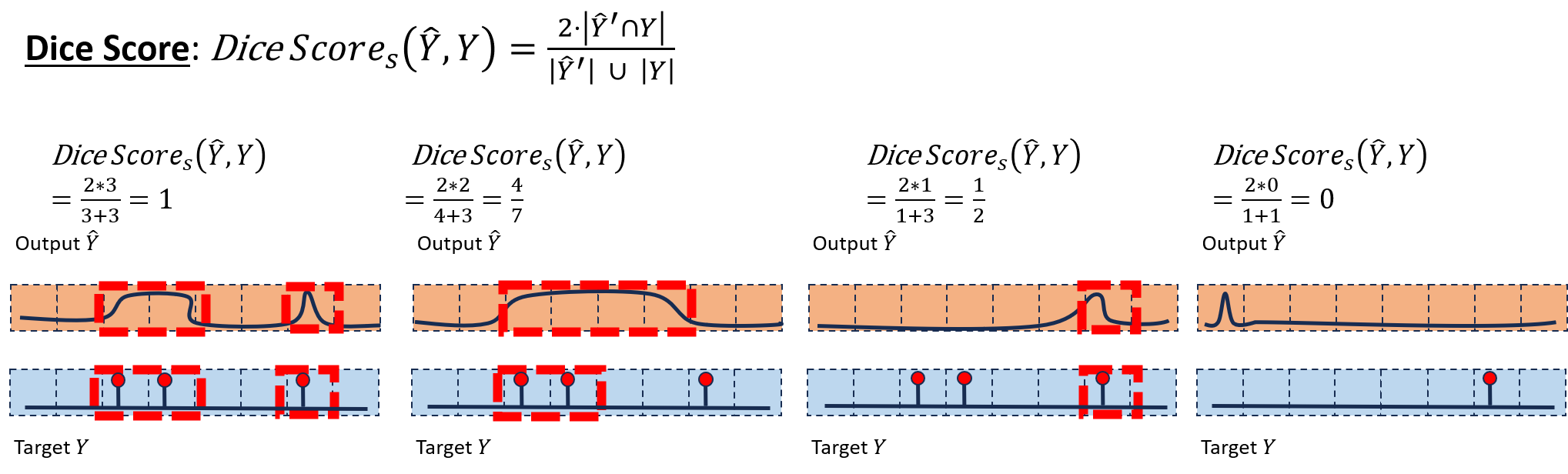}
        \caption{Dice Score}
        \label{fig:anomaly_dicescore}
    \end{subfigure}
    
    \caption{Anomaly Prediction Metrics}
    \label{fig:anomaly_prediction_metrics}
\end{figure}

To evaluate the quality of anomaly predictions, we use the following metrics (Figure \ref{fig:anomaly_prediction_metrics}):

\subsubsection{  \textbf{Existence of Anomaly}  }
   This metric evaluates whether the model correctly predicts the existence of at least one anomaly within the prediction range. It is a binary measure that compares the presence of any anomaly in the predicted range with the ground truth. We define the True Positive (TP), False Positive (FP), False Negative (FN) for each pair of predicted probability segment of anomaly $\hat{Y}_t$ and labels $Y_t$ as follows:
   \begin{align}
   \text{TP}(\hat{Y}_t, Y_t) = \begin{cases} 
   1 & \text{if } Sum(\hat{Y}_t) \geq s \text{ and } Sum(Y_t) \geq s\\
   0 & \text{otherwise} 
   \end{cases}
   \end{align}
   \begin{align}
   \text{FP}(\hat{Y}_t, Y_t) = \begin{cases} 
   1 & \text{if } Sum(\hat{Y}_t) \geq s \text{ and } Sum(Y_t) < s\\
   0 & \text{otherwise} 
   \end{cases}
   \end{align}
   \begin{align}
   \text{FN}(\hat{Y}_t, Y_t) = \begin{cases} 
   1 & \text{if } Sum(\hat{Y_t}) < s \text{ and } Sum(Y_t) \geq s\\
   0 & \text{otherwise} 
   \end{cases}
   \end{align}
   where $Sum(Y_t)=\sum_{i=1}^{T} y_{t+i} $ and $s \in [0, 1)$ is a threshold. 
   
   The metric \textbf{Existence} of anomaly that measures the overall performance on the total pairs ($\hat{Y}$, $Y$) is defined as:
    \begin{align}
   \text{Exist}(\hat{Y}, Y) = \frac{2 \cdot TP}{2 \cdot TP + FP + FN}
    \end{align}
   where TP, FP, and FN are the sum value of all pairs of predicted probability segments of anomaly $\hat{Y}$ and labels $Y$ over the total timestamps. 
   
   This metric provides a simple outcome indicating how well the model correctly identified the presence of any anomaly in the prediction interval.

\subsubsection{ \textbf{Density of Anomalies}}
   This metric measures the difference in the cumulative densities of predicted anomalies compared to the actual count of anomaly labels. It's calculated as:
   \begin{align}
   \text{Density}(\hat{Y}_t, Y_t) = 1 - \frac{1}{T}\left| \sum_{i=1}^{T} (\hat{y}_{t+i} -  y_{t+i} )\right|
   \end{align}
   This metric evaluates the discrepancy in the cumulative density of predicted anomalies versus the count of actual anomaly labels. This metric ranges from $0$ to $1$, where $1$ indicates that the size of the predicted anomaly is exact, and values closer to $0$ indicate over-predicting or under-predicting the size of anomaly events.

\subsubsection{\textbf{Lead Time}}

   Lead time measures how close the first predicted anomaly is to the first observed ground truth anomaly. It is calculated as:
  \begin{align}
   \text{LeadTime}(\hat{Y}_t, Y_t) = 1 - \frac{1}{T}\left|i_{\hat{y}} - i_y\right|
  \end{align}
   where  $ i_y = \min \{i \mid y_{t+i} = 1\} $ is the index of the first observed ground truth anomaly, 
    $ i_{\hat{y}} = \min \{i \mid \hat{y}_{t+i} \geq s\} $ is the index of the first predicted anomaly probabilities that are greater than user-defined threshold $ s $.

   This metric ranges from $0$ to $1$, where $1$ indicates that the first predicted anomaly exactly matches the timing of the first ground truth anomaly, and values closer to 0 indicate a larger discrepancy between the predicted and actual timing.

\subsubsection{\textbf{Dice Score }}

   The Dice Score, also known as the F1 Score for sets, measures the overlap between the predicted anomalies and the ground truth anomalies. It is defined as:
   \begin{align}
   \text{DiceScore}( \hat{Y}_t, Y_t) = \frac{2  |\hat{Y}_t' \cap Y_t|}{|\hat{Y}_t'| + |Y_t|}
   \end{align}
   where $\hat{Y}_t'= \mathbbm{1}(\hat{Y}_t \geq s)$ is an indicator function that converts the probabilities to booleans. This metric ranges from $0$ to $1$, where $1$ indicates perfect coverness between the predicted anomalies $\hat{y}$ and the ground truth $y$, and $0$ indicates no overlap.

Using these metrics, we can comprehensively evaluate the performance of the anomaly prediction model, considering both the accuracy of the predicted anomaly intervals and the timeliness of the predictions.

\subsection{Models}
In this section, we describe the models employed in our experiments for anomaly detection in time series data. Our selection includes some of the most recent and fundamental techniques: Fully Connected Network, PatchTST, and Kernel U-Net. These models were chosen for their simplicity and demonstrated performance in the field of time series forecasting.

\textbf{Fully Connected Network} (FCN) serves as a baseline model for anomaly prediction in time series data. This model applies multiple linear layers with ReLU activation functions to extract temporal patterns. Due to its simplicity and effectiveness, we use it as a baseline to compare the performance of more complex models in time series anomaly prediction.

\textbf{PatchTST} (Patch Time Series Transformer) \cite{Nie-2023-PatchTST} is a novel model designed to capture long-term dependencies in time series data through the use of patches, analogous to those used in computer vision transformers. The core idea of PatchTST is to segment the time series into patches and process them using transformer encoders. This approach relieves the overfitting of the transformer and reduces the complexity to $O(\frac{L}{S}log(\frac{L}{S}))$, where $L$ is the length and $S$ is patch size.  

\textbf{ Kernel U-Net} 
 \cite{you_kun_2024} is an extension of the traditional U-Net architecture, specifically designed to address the challenges in time series forecasting. This model separates the process of partitioning input time series into patches from kernel manipulation, allowing for customized kernel execution. Kernel U-Net keeps a symmetric, hierarchical U-shaped neural network structure, which ensures its linear computation complexity $O(L)$, and makes it more suitable for diverse and large-scale time series applications.

\section{Dataset}
Real-world datasets for time series anomaly prediction can be scarce, incomplete, or lack sufficient anomalies to train and evaluate models effectively. This limitation necessitates the creation of synthetic datasets that can simulate various scenarios of anomalies. Synthetic datasets allow researchers to control the parameters and conditions under which anomalies occur, providing a robust platform for benchmarking and validating anomaly prediction models.
In this section, we describe a synthetic dataset, a private dataset designed for anomaly prediction, and 3 datasets from anomaly detection. The descriptions of datasets are in Table \ref{table:dataset_description}.

\begin{figure}[t!]
    \centering
    \includegraphics[width=\linewidth]{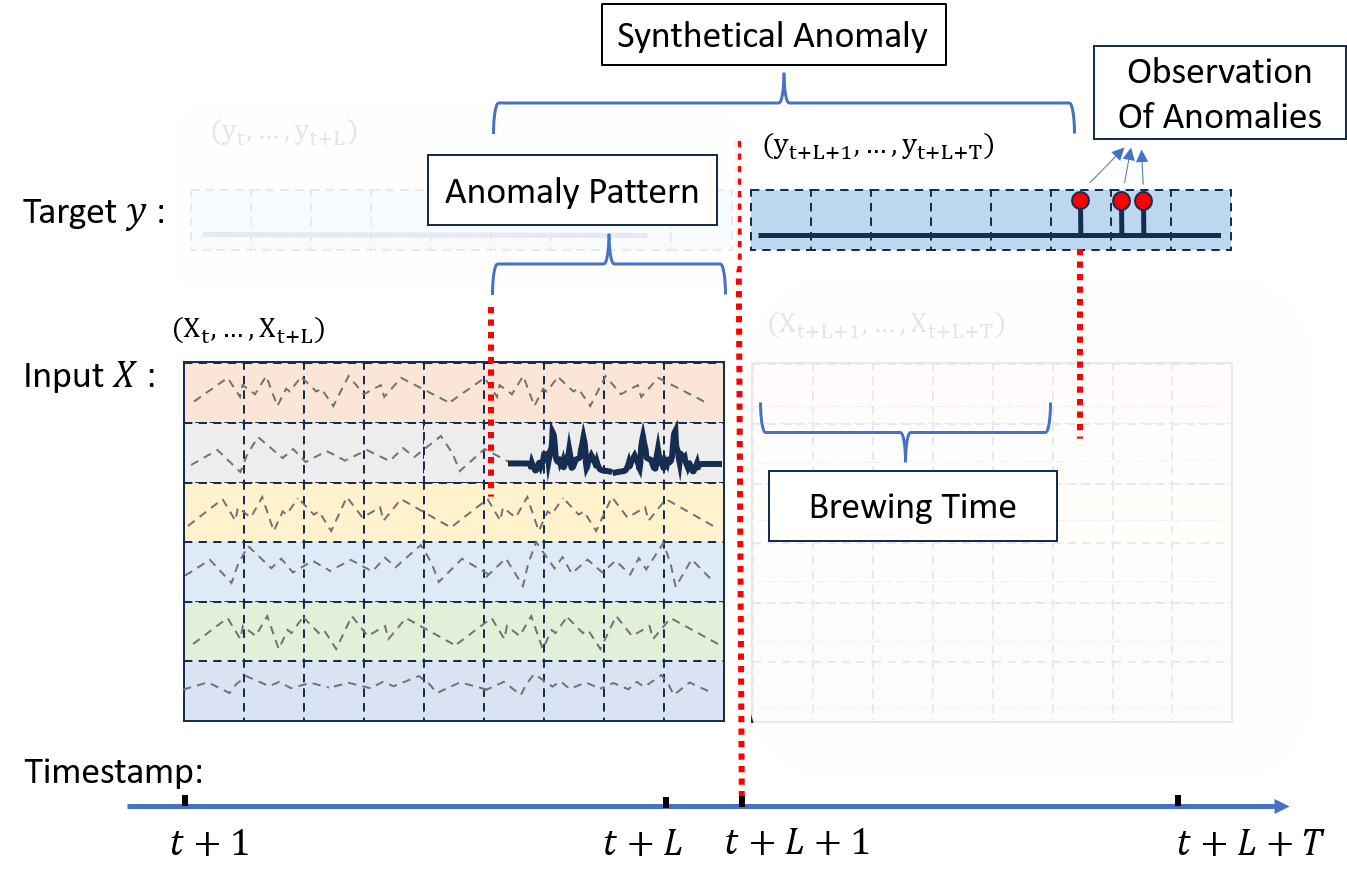}
    \caption{Synthetic Anomaly Example}
    \label{fig:synthetical_anomaly}
\end{figure}

\subsection{Synthetic Anomaly Prediction Dataset}

The synthetic datasets are designed to evaluate and benchmark anomaly prediction algorithms. Datasets 1-5 are univariate, while datasets 6-10 are multivariate. The time series patterns include sinusoidal and cosine waves, with configurations being either fixed or mixed. Both the brewing time and anomaly labels are either fixed or follow a Gaussian distribution. The difficulty of the datasets increases progressively. 

\subsubsection{\textbf{Synthetic Time Series Anomaly}}

A synthetic anomaly in a time series is defined by a time series pattern, a segment that includes an anomaly pattern, brewing time, and a sequence of positive labels at the observation period (Figure \ref{fig:synthetical_anomaly}). The key components are:
\begin{itemize}
  
  \item  \textbf{Time Series Pattern}: A series of points generated by sine or cosine function. We use a single fixed sine function (Fixed), a mixed sine function of different frequencies (Mixed), multivariate series composed of sine or cosine functions (Mixed).
  \item  \textbf{Anomaly Pattern}: A pattern that the model can detect. It is usually not observable by human experts but contains decisive patterns indicative of an anomaly in the future. We use a fixed pattern composed of a segment of Gaussian random points and an increasing line (Fixed). 
 
  \item \textbf{Brewing Time}: The interval between the end of the hidden anomaly pattern and the observable anomalies annotation. We use a fixed number(Fixed) or a random number following Gaussian distribution (Gaussian).

   \item \textbf{Observation}: A sequence of annotation of anomaly after the brewing time. It is typically visually evident to human experts, prompting them to raise an alert. The anomaly event is labeled as a positive example during the observation. We use a fixed number (Fixed) or a random number (Gaussian) for the size of anomaly labels.

\end{itemize}

 Table \ref{table:synthetic_datasets_description} resumes the details of the generated synthetic dataset for anomaly prediction. 

\subsubsection{\textbf{Signal-to-Noise Ratio (SNR)}} 
SNR is crucial in communication and can also be applied to anomaly detection. In this context, SNR is the user-defined ratio of the weighted sum of the inserted anomaly pattern and original time series. SNR defines the separability of anomaly and normal segments and quantifies the information in the anomaly pattern.

 We note the \textbf{Signal} as an important anomaly pattern detected by a model. We note the \textbf{Noise} as an irrelevant pattern received in a model. We note \textbf{Signal-to-Noise Ratio (SNR)} as the fraction of amplitude of anomaly pattern to that of irrelevant pattern. A higher SNR indicates that the signal is evident, while a lower SNR means the signal is not evident.

\begin{table}[t!]
\setlength{\tabcolsep}{3.5pt} 

\centering
\caption{Dataset Description for Synthetic, SBDA, MSL, SMAP, and SMD}
\begin{tabular}{lccccc}
\toprule
\textbf{Dataset} & \begin{tabular}[c]{@{}c@{}}\textbf{Dataset }\\\textbf{~Points}\end{tabular} & \begin{tabular}[c]{@{}c@{}}\textbf{Anomaly }\\\textbf{~Labels}\end{tabular}  & \begin{tabular}[c]{@{}c@{}}\textbf{Anomaly }\\\textbf{~Ratio}\end{tabular}  & \begin{tabular}[c]{@{}c@{}}\textbf{Features }\\\textbf{Size}\end{tabular} & \textbf{Instances} \\
\midrule
Synthetic (Avg) & 10,000  & 100 & 1.00\%  & 1 and 3 & 10\\
SBDA (Avg)  & 5,921 & 123 & 2.08\% & 52 & 10\\
MSL & 73,729 & 7,766  & 10.53\% & 55 & 1\\
SMAP  & 427,617  & 54,696 & 12.79\% & 25 & 1\\
SMD (Avg) & 25,300  & 1,051 & 4.15\% & 38 & 28\\
\bottomrule
\end{tabular} 

\label{table:dataset_description}
\end{table}

\begin{table}[t!]
\centering
\caption{Description of 10 synthetic datasets}
\begin{tabular}{lccccc}
\toprule
\textbf{Dataset} & \begin{tabular}[c]{@{}c@{}}\textbf{Time Series }\\\textbf{~Pattern}\end{tabular} & \begin{tabular}[c]{@{}c@{}}\textbf{Anomaly }\\\textbf{Pattern}\end{tabular} & \begin{tabular}[c]{@{}c@{}}\textbf{Brewing }\\\textbf{Time}\end{tabular} & \begin{tabular}[c]{@{}c@{}}\textbf{Anomaly}\\\textbf{~Label}\end{tabular}  \\ 
\midrule

Synthetic\_1 & Fixed  & Fixed & Fixed  & Fixed \\ 
Synthetic\_2 & Fixed  & Fixed & Gaussian & Fixed \\ 
Synthetic\_3 & Fixed  & Fixed & Fixed  & Gaussian  \\ 
Synthetic\_4 & Fixed  & Fixed & Gaussian & Gaussian  \\ 
Synthetic\_5 & Mixed  & Fixed & Gaussian & Gaussian  \\ 
Synthetic\_6 & Multi-Fixed  & Fixed & Fixed  & Fixed \\ 
Synthetic\_7 & Multi-Fixed  & Fixed & Gaussian & Fixed \\ 
Synthetic\_8 & Multi-Fixed  & Fixed & Fixed  & Gaussian  \\ 
Synthetic\_9 & Multi-Fixed  & Fixed & Gaussian & Gaussian  \\ 
Synthetic\_10  & Multi-Mixed  & Fixed & Gaussian & Gaussian  \\ 
\bottomrule
\end{tabular}
\label{table:synthetic_datasets_description}
\end{table}

\begin{table*}[tp!]
\setlength{\tabcolsep}{5.5pt} 

\centering
\tiny
\caption{Performance Scores for Different Models on Synthetic Datasets 1-10}
\label{tab:scores}
\begin{tabular}{ccccccccccccccccccc}
\toprule
\textbf{Dataset} & \multicolumn{3}{c}{\textbf{Wasserstein} $\downarrow$} & \multicolumn{3}{c}{\textbf{Existence} $\uparrow$} & \multicolumn{3}{c}{\textbf{Density Sum} $\uparrow$} & \multicolumn{3}{c}{\textbf{Lead Time} $\uparrow$} & \multicolumn{3}{c}{\textbf{Dice Score} $\uparrow$} \\
 \cmidrule(lr){2-4} \cmidrule(lr){5-7} \cmidrule(lr){8-10} \cmidrule(lr){11-13} \cmidrule(lr){14-16}
 \cmidrule(lr){17-19}
 & \textbf{ FCN} & 
 \textbf{PatchTST} &
 \textbf{K-U-Net} & \textbf{ FCN} & 
 \textbf{PatchTST} & 
 \textbf{K-U-Net} & \textbf{ FCN} & \textbf{PatchTST} & 
 \textbf{K-U-Net} & \textbf{ FCN} & \textbf{PatchTST} & 
 \textbf{K-U-Net} & \textbf{ FCN} & \textbf{PatchTST} & 
 \textbf{K-U-Net} \\
 \midrule
 Synthetic\_1 & 0.003 & 0.003 & 0.004 & 0.956 & 0.643 & 0.958 & 0.995 & 0.994 & 0.996 & 0.936 & 0.852 & 0.953 & 0.391 & 0.272 & 0.369 \\
Synthetic\_2 & 0.004 & 0.005 & 0.004 & 0.783 & 0.532 & 0.873 & 0.995 & 0.995 & 0.995 & 0.879 & 0.847 & 0.884 & 0.131 & 0.093 & 0.143 \\
Synthetic\_3 & 0.013 & 0.014 & 0.012 & 0.956 & 0.898 & 0.963 & 0.991 & 0.990 & 0.991 & 0.925 & 0.904 & 0.935 & 0.700 & 0.669 & 0.710 \\
Synthetic\_4 & 0.019 & 0.024 & 0.018 & 0.901 & 0.772 & 0.906 & 0.987 & 0.987 & 0.988 & 0.885 & 0.853 & 0.904 & 0.588 & 0.566 & 0.588 \\
Synthetic\_5 & 0.027 & 0.026 & 0.028 & 0.897 & 0.788 & 0.891 & 0.986 & 0.986 & 0.986 & 0.808 & 0.795 & 0.827 & 0.508 & 0.498 & 0.517 \\
Synthetic\_6 & 0.002 & 0.003 & 0.002 & 0.935 & 0.258 & 0.960 & 0.996 & 0.995 & 0.996 & 0.943 & 0.840 & 0.950 & 0.406 & 0.115 & 0.459 \\
Synthetic\_7 & 0.004 & 0.004 & 0.003 & 0.785 & 0.024 & 0.852 & 0.995 & 0.996 & 0.995 & 0.833 & 0.850 & 0.868 & 0.148 & 0.000 & 0.173 \\
Synthetic\_8 & 0.028 & 0.023 & 0.024 & 0.912 & 0.469 & 0.947 & 0.988 & 0.985 & 0.990 & 0.911 & 0.785 & 0.927 & 0.676 & 0.375 & 0.692 \\
Synthetic\_9 & 0.046 & 0.046 & 0.042 & 0.826 & 0.305 & 0.838 & 0.985 & 0.982 & 0.985 & 0.805 & 0.720 & 0.807 & 0.507 & 0.392 & 0.516 \\
Synthetic\_10 & 0.047 & 0.051 & 0.048 & 0.465 & 0.149 & 0.786 & 0.982 & 0.986 & 0.985 & 0.815 & 0.514 & 0.823 & 0.392 & 0.062 & 0.482 \\
SBAD & 0.036 & 0.045 & 0.037 & 0.398 & 0.396 & 0.352 & 0.997 & 0.995 & 0.997 & 0.741 & 0.754 & 0.739 & 0.158 & 0.115 & 0.182 \\
MSL & 0.350 & 0.245 & 0.133 & 0.197 & 0.213 & 0.190 & 0.933 & 0.988 & 0.992 & 0.810 & 0.846 & 0.882 & 0.519 & 0.562 & 0.639 \\
SMAP & 0.487 & 0.622 & 0.511 & 0.232 & 0.235 & 0.245 & 0.990 & 0.989 & 0.989 & 0.927 & 0.925 & 0.709 & 0.802 & 0.743 & 0.513 \\
SMD & 0.083 & 0.076 & 0.091 & 0.198 & 0.126 & 0.216 & 0.991 & 0.990 & 0.992 & 0.839 & 0.840 & 0.850 & 0.439 & 0.353 & 0.445 \\ 
\bottomrule 
\end{tabular}
\end{table*}

\subsection{Anomaly Detection Dataset}

\subsubsection{\textbf{SMD}} The Server Machine Dataset (SMD) \footnote{https://github.com/NetManAIOps/OmniAnomaly} \cite{omnianomaly_2019} is a widely utilized dataset for anomaly detection in time series. It contains data collected from 28 servers and includes a mix of normal and abnormal behaviors. The dataset is typically used to evaluate the performance of unsupervised anomaly detection algorithms in industrial settings. We use the test set of SMD for evaluating our proposed supervised anomaly prediction task.

\subsubsection{\textbf{SBAD}} The Server Breakdown Anomaly Detection (SBAD) dataset is a private anonymized dataset. It is composed of 16 metrics, such as CPU, memory usage, and other system-related measurements, and 36 label columns annotated by experts. It contains 10 data instances collected over six months, capturing normal operations and server failures. This dataset is instrumental for evaluating and developing anomaly detection algorithms to identify server breakdowns. Compared with the SMD dataset, the SBAD dataset contains not only physical metrics but also additional expert knowledge through annotations.

\subsubsection{\textbf{MSL}} The Mars Science Laboratory (MSL) \footnote{https://pds-imaging.jpl.nasa.gov/volumes/msl.html} dataset originates from the telemetry data of NASA's Curiosity rover. It includes multiple telemetry channels recorded during the rover's mission on Mars. This dataset is often used in research to develop and test algorithms for anomaly detection in space mission data. We only use the test set of MSL for evaluation. 

\subsubsection{\textbf{SMAP}} The Soil Moisture Active Passive (SMAP) \footnote{https://smap.jpl.nasa.gov/data/} dataset is derived from the NASA satellite mission aimed at measuring soil moisture levels. It encompasses a variety of telemetry channels that capture the satellite's operational data. Researchers utilize this dataset to benchmark and improve anomaly detection methods for satellite telemetry. We only use the test set of SMAP for evaluation.

\section{Experiments and Results}
In this section, we present the experiments conducted to evaluate the performance of anomaly prediction models using both synthetic and real-world datasets. The datasets are split into training, validation, and test sets with a ratio of 0.7, 0.1, and 0.2, respectively, with anomalies distributed across each subset in the same ratio.  For all datasets, the input length is set to 50, while the output length is 20 for the synthetic dataset and 50 for the real-world datasets. All experiments use a learning rate of 0.0005 and a hidden dimension of 128. For the models PatchTST and K-U-Net, a patch size of 5 is employed, with K-U-Net utilizing an MLP kernel, and both models consisting of a single layer. The training is conducted for up to 100 epochs, and patience for early stopping is set to 10 epochs. The threshold $s$ is empirically set to be 0.1.

The experimental results demonstrated the high accuracy of the models in detecting the presence of anomalies across different synthetic datasets (Table \ref{tab:scores}). FCN and K-U-Net correctly identified anomalies in over 90\% of the test cases in a univariate setting, as indicated by the existence of anomaly metric. However, PatchTST quickly overfitted to particular cases, achieving only a 75\% F1-score. We measured the Density, Lead Time, and Dice Score in cases where the models successfully predicted the existence of an anomaly (True Positives). The density sum showed that the models' predictions closely matched the total anomalies expected in the future horizon, with all models reaching 0.95. The lead time metric indicated that the models' predictions were closely aligned with the actual occurrences of anomalies, with an average lead time score of 0.85, suggesting effective prediction with minimal delay. The Dice score averaged below 0.5 due to the challenges posed by temporal dynamics, reflecting that the overlap between predicted and actual anomalies varied depending on the distribution of anomaly labels and brewing time. Results for the existence of anomaly on real-world datasets such as SBAD, SMD, SMAP, and MSL were generally lower than 0.5.

\begin{figure}[t!]
    \centering
    \includegraphics[width=1.0\linewidth]{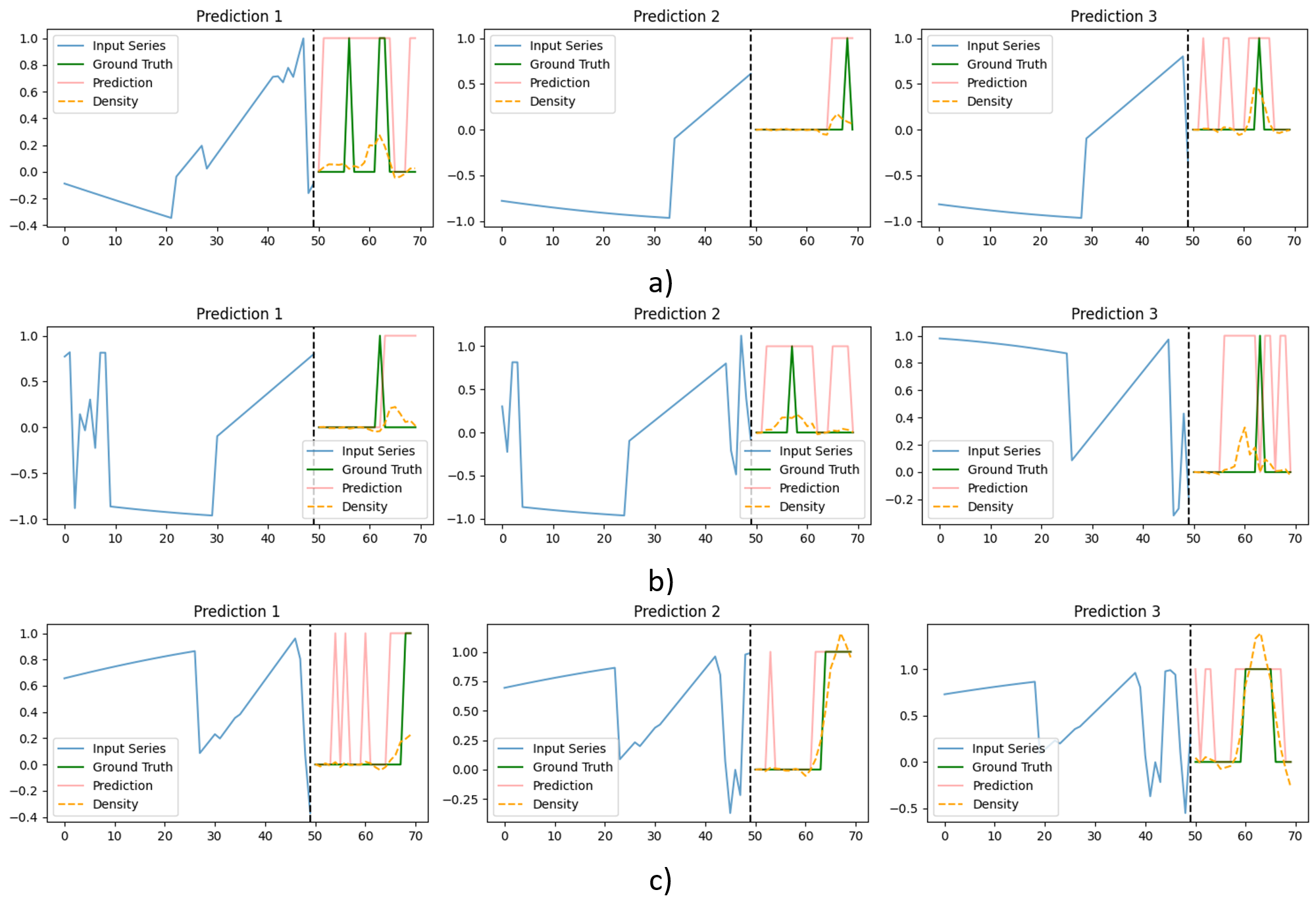}
    \caption{Anomaly Prediction on Univariate Synthetic Dataset}
    \label{fig:univariate_synthetic_dataset}
\end{figure}
\begin{figure}[t!]
    \centering
    \includegraphics[width=1.0\linewidth]{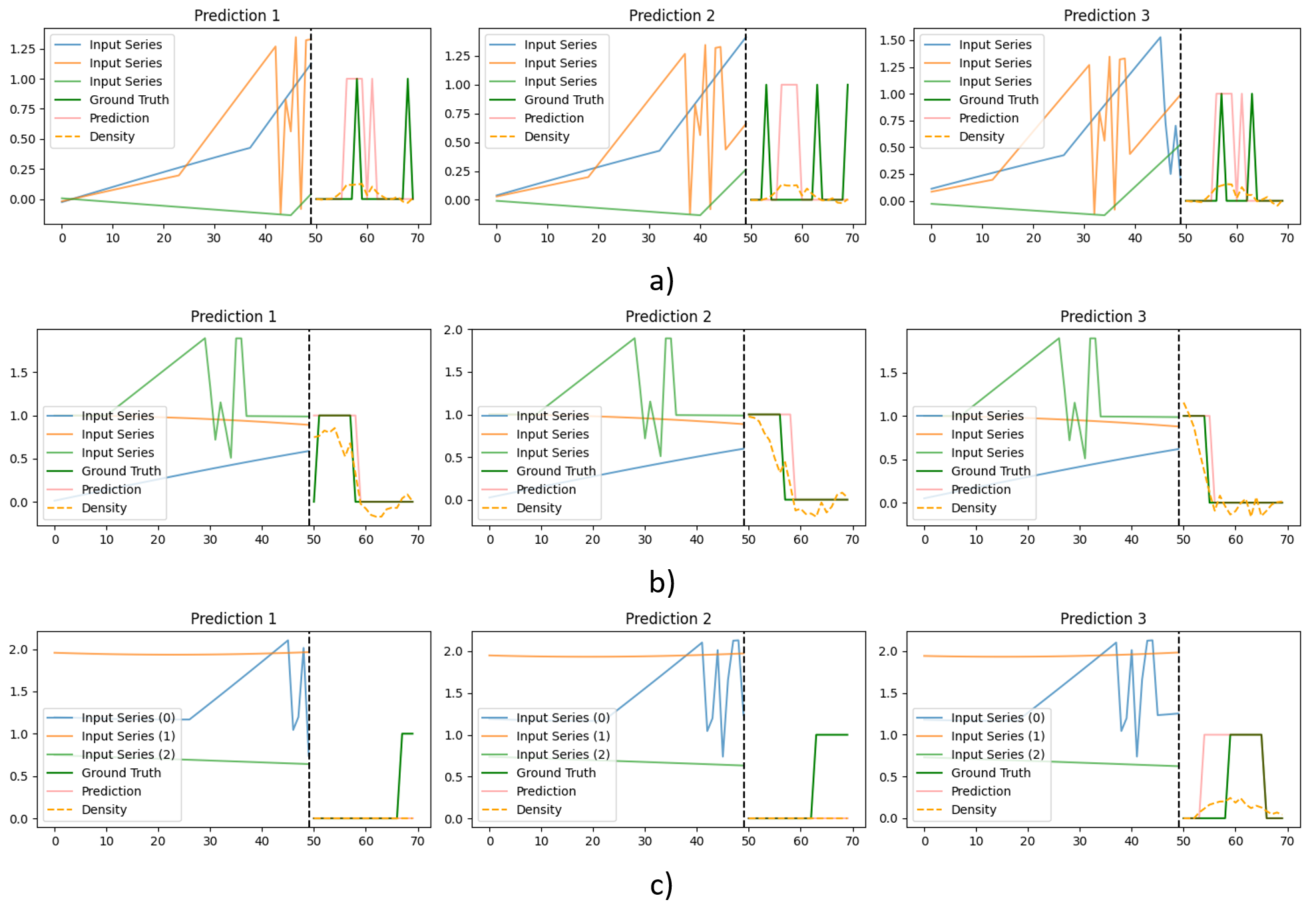}
    \caption{Anomaly Prediction on Multivariate Synthetic Dataset}
    \label{fig:multivariate_synthetic_dataset}
\end{figure}

Overall, these results provide proof of concept for our anomaly prediction approach. The use of synthetic datasets allowed us to rigorously test the models under controlled conditions, ensuring their reliability before applying them to real-world applications. The results on real-world datasets initially validate the concept but also reveal existing challenges in achieving high accuracy. We share examples of anomaly prediction on synthetic datasets in Figures \ref{fig:univariate_synthetic_dataset}, \ref{fig:multivariate_synthetic_dataset}.

\section{Conclusion}
In this work, we introduce anomaly prediction, a novel approach that integrates delay and horizon into classic anomaly detection, providing probability densities for identifying potential anomalies in a future horizon. By leveraging state-of-the-art time series forecasting techniques and anomaly detection, we validate our approach on a synthetic dataset specifically designed for benchmarking anomaly prediction models. We also apply our method to real-world datasets, compiling the first benchmark for anomaly prediction, which offers valuable insights into its practical performance. 

This work advances the field of anomaly detection by proposing a novel approach to anomaly prediction, laying a solid foundation for future research and practical applications. Future work may focus on enhancing the model's adaptability and precision, addressing the identified challenges such as accuracy and temporal dynamics, and developing more reliable anomaly prediction tools for diverse real-world applications.


\bibliographystyle{nips.bst}
\bibliography{kun.bib}

\end{document}